\documentclass[11pt,a4paper]{article}
\usepackage[hyperref]{emnlp2018}
\usepackage{times}
\usepackage{latexsym}
\usepackage{graphicx}
\usepackage{amsmath,amssymb}
\usepackage{algorithm,algorithmic}
\usepackage{arydshln}
\usepackage{subcaption}
\usepackage{xspace}

\usepackage{mathtools}
\usepackage{tikz}
\usetikzlibrary{calc,trees,positioning,arrows,chains,shapes.geometric,%
    decorations.pathreplacing,decorations.pathmorphing,shapes,%
    matrix,shapes.symbols}

\tikzset{suppress join/.code={\def\tikz@after@path{}}}

\tikzset{
>=stealth',
  punktchain/.style={
    rectangle, 
    rounded corners, 
    fill=blue!10,
    draw=black, thin,
    text width=6em, 
    minimum height=3em, 
    text centered,
    on chain},
  circlechain/.style={
    circle,  
    draw=black, thin,
    text centered, 
    minimum size=3em,
    inner sep=1pt,
    on chain},
  gen_circle/.style={
    circle,  
    draw=black, thin,
    text centered, 
    minimum size=3em,
    inner sep=1pt,
    on chain},
  infer_circle/.style={
    circle,  
    draw=black, thin,
    text centered, 
    minimum size=3em,
    inner sep=1pt,
    suppress join,
    join=by {-latex', dashed}
    },
  punktchain2/.style={
    rectangle,
    dashed,
    rounded corners, 
    draw=black, thin,
    text width=6em, 
    minimum height=5em, 
    text centered, 
    on chain},
  line/.style={draw, thick, <-},
  side/.style={
    rectangle,
    minimum width=1em,
    draw=black, thin,
    text width=2em, 
    minimum height=2em, 
    text centered},
  sidecircle/.style={
    circle,  
    fill=green!10,
    draw=black, thin,
    text centered, 
    minimum size=0.1em,
    inner sep=1pt,
    on chain},
  every join/.style={->, thin,shorten >=1pt},
  decoration={brace},
  tuborg/.style={decorate},
  tubnode/.style={midway, right=2pt},
  bignode/.style={font={\fontsize{18}{22.4}\selectfont}},
  smallnode/.style={font={\fontsize{10}{12}\selectfont}},
}

\usepackage{url}

\DeclareCaptionFont{tenpt}{\fontsize{10pt}{11pt}\selectfont#1}
\aclfinalcopy 

\newcommand{\mccomment}[1]{\textcolor{red}{\bf \small [ #1 --MC]}}
\newcommand{\qtcomment}[1]{\textcolor{blue}{\bf \small [ #1 --QT]}}
\newcommand{\kgcomment}[1]{\textcolor{orange}{\bf \small [ #1 --KG]}}
\newcommand{\klcomment}[1]{\textcolor{magenta}{\bf \small [ #1 --KL]}}

\newcommand{\qtcomments}[1]{\textcolor{gray}{\bf \small [ #1 --QT]}}
\newcommand{\kgcomments}[1]{\textcolor{gray}{\bf \small [ #1 --KG]}}
\newcommand{\klcomments}[1]{\textcolor{gray}{\bf \small [ #1 --KL]}}
\renewcommand{\mccomment}[1]{}
\renewcommand{\qtcomment}[1]{}
\renewcommand{\kgcomment}[1]{}
\renewcommand{\klcomment}[1]{}
\renewcommand{\qtcomments}[1]{}
\renewcommand{\kgcomments}[1]{}
\renewcommand{\klcomments}[1]{}
\newcommand{\vsmg}{VSL-G\xspace}
\newcommand{\vsmggflat}{VSL-GG-Flat\xspace}
\newcommand{\vsmgghier}{VSL-GG-Hier\xspace}

\newcommand{\kld}{\mathit{KL}}

\title{Variational Sequential Labelers for Semi-Supervised Learning}
\author{Mingda Chen\qquad Qingming Tang\qquad Karen Livescu\qquad Kevin Gimpel\\
Toyota Technological Institute at Chicago, Chicago, IL, 60637, USA\\
  {\tt \{mchen,qmtang,klivescu,kgimpel\}@ttic.edu}\\}

\date{}

\begin{document}
\maketitle
\begin{abstract}
We introduce a family of multitask variational methods for semi-supervised sequence labeling. Our model family consists of a latent-variable generative model and a discriminative labeler.
The generative models use latent variables to define the conditional probability of a word given its context, drawing inspiration from word prediction objectives commonly used in learning word embeddings.
The labeler
helps inject discriminative information into the latent space.
We explore several latent variable configurations, including ones with hierarchical structure, which enables the model to account for both label-specific and word-specific information.
Our models consistently outperform standard sequential baselines on 8 sequence labeling datasets, and improve further with unlabeled data.
\end{abstract}

\section{Introduction}

Sequence labeling tasks in natural language processing (NLP) often have limited annotated data available for model training. In such cases regularization can be important, and it can be helpful to use additional unlabeled data. One approach for both regularization and semi-supervised training is to design latent-variable generative models and then develop neural variational methods for learning and inference~\cite{kingma2013auto,rezende2015variational}.

Neural variational methods have been quite successful for both generative modeling and representation learning, and have recently been applied to a variety of NLP tasks~\cite{mnih2014neural,bowman2016generating,miao2016neural,serban2017piecewise,zhou2017multi,hu2017toward}. They are also very popular for semi-supervised training; when used in such scenarios, they typically have an additional task-specific prediction loss~\cite{kingma2014semi,maaloe2016auxiliary,zhou2017multi,yang2017improved}. However, it is still unclear how to use such methods in the context of sequence labeling.

In this paper, we apply neural variational methods to sequence labeling by combining a latent-variable generative model and a discriminatively-trained labeler.
We refer to this family of procedures as variational sequential labelers (VSLs).
Learning  maximizes the conditional probability of
each word given its context and minimizes the classification loss given the latent space.
We explore several models within this family that use different kinds of conditional independence structure among the latent variables within each time step. Intuitively, the multiple latent variables can disentangle information pertaining to label-oriented and word-specific properties.

We study VSLs in the context of named entity recognition (NER) and several part-of-speech (POS) tagging tasks, both on English Twitter data and on data from six additional languages.
Without unlabeled data, our models consistently show 0.5-0.8\% accuracy improvements across tagging datasets and 0.8 $F_1$ improvement for NER. Adding unlabeled data further improves the model performance by 0.1-0.3\% accuracy or 0.2 $F_1$ score.
We obtain the best results with a hierarchical structure using two latent variables at each time step.

Our models, like generative latent variable models in general, have the ability to naturally combine labeled and unlabeled data. We obtain small but consistent performance improvements by adding unlabeled data.  In the absence of unlabeled data, the variational loss
acts as regularizer on the learned representation of the supervised sequence prediction model.
Our results demonstrate that this regularization improves performance
even when only labeled data is used.
We also compare different ways of applying the classification loss when using a latent variable hierarchy, and find that the most effective structure also provides the cleanest separation of information in the latent space.

\section{Related Work}

There is a growing amount of work applying neural variational methods to NLP tasks, including document modeling~\cite{mnih2014neural,miao2016neural,serban2017piecewise}, machine translation~\cite{zhang2016variational}, text generation~\cite{bowman2016generating,serban2017piecewise,hu2017toward}, language modeling~\cite{bowman2016generating,yang2017improved}, and sequence transduction~\cite{zhou2017multi}, but we are not aware of any such work for sequence labeling. Before the advent of neural variational methods, there were several efforts in latent variable modeling for sequence labeling~\cite{quattoni2007hidden,SunT09}.

There has been a great deal of work on using variational autoencoders in semi-supervised settings~\cite{kingma2014semi,maaloe2016auxiliary,zhou2017multi,yang2017improved}.
Semi-supervised sequence labeling has a rich history~\cite{altun2006maximum,jiao-EtAl:2006:COLACL,mann-mccallum:2008:ACLMain,subramanya-petrov-pereira:2010:EMNLP,sogaard:2011:ACL-HLT20111}. The simplest methods, which are also popular currently, use representations learned from large amounts of unlabeled data~\cite{miller-guinness-zamanian:2004:HLTNAACL,owoputi-13, peters2017semi}. Recently,~\citet{zhang2017semi} proposed a structured neural autoencoder that can be jointly trained on both labeled and unlabeled data.

Our work involves multi-task losses and is therefore also related to the rich literature on
multi-task learning for sequence labeling~\cite[\emph{inter alia}]{plank2016multilingual,augenstein2017multi,bingel2017identifying,rei2017semi}.

Another related thread of work is learning interpretable latent representations. \citet{zhou2017multi} factorize an inflected word into lemma and morphology labels, using continuous and categorical latent variables. \citet{hu2017toward} interpret a sentence as a combination of an unstructured latent code and a structured latent code, which can represent attributes of the sentence.

There have been several efforts in combining variational autoencoders and recurrent networks~\cite{gregor2015draw,chung2015recurrent,fraccaro2016sequential}. While the details vary, these models typically contain latent variables at each time step in a sequence. This prior work mainly focused on ways of parameterizing the time dependence between the latent variables, which gives them more power in modeling distributions over observation sequences. In this paper, we similarly use latent variables at each time step, but we adopt stronger independence assumptions which leads to simpler models and inference procedures. Also, the models cited above were developed for modeling data distributions, rather than for supervised or semi-supervised learning, which is our focus here.

\klcomment{shortened this par as it seemed to be veering away from related work a bit.  please make sure I haven't removed anything important.}
The key novelties in
our work compared to the prior work mentioned above
are the proposed sequential variational labelers and the investigation of latent variable hierarchies within these models.  The empirical effectiveness of latent hierarchical structure in variational modeling is a key contribution of this paper and may be applicable to the other applications discussed above.  Recent work, contemporaneous with this submission, similarly showed the advantages of combining hierarchical latent variables and variational learning for conversational modeling, in the context of a non-sequential model~\cite{VHCR:2018:NAACL}.
\section{Proposed Methods}

We begin by describing variational autoencoders and the notation we will use in the following sections. We denote the input word sequence by $x_{1:T}$, the corresponding label sequence by $l_{1:T}$, the input words other than the word at position $t$ by $x_{-t}$, the generative model by $p_\theta(\cdot)$, and the posterior inference model by $q_\phi(\cdot)$.

\subsection{Background: Variational Autoencoders}
We review variational autoencoders (VAEs) by  describing a VAE for an input sequence $x_{1:T}$. When using a VAE, we assume a generative model that generates an input using a latent variable $z$, typically assumed to follow a multivariate Gaussian distribution. We seek to maximize the marginal likelihood of inputs $x_{1:T}$ when marginalizing out the latent variable $z$. Since this is typically intractable, especially when using continuous latent variables, we instead maximize a lower bound on the marginal log-likelihood~\citep{kingma2013auto}:

\begin{equation}
\begin{aligned}
    &\log~p_\theta(x_{1:T})\geq\\
    &\mathop\mathbb{E}_{z\sim q_\phi(\cdot\vert x_{1:T})}\!\!\left[\log p_\theta(x_{1:T}\vert z)-\log\frac{q_\phi(z\vert~x_{1:T})}{p_\theta(z)} \right]=\\
    &\underbrace{\mathop\mathbb{E}_{z\sim q_\phi(\cdot\vert x_{1:T})}\!\!\!\!\!\!\!\!\left[\log p_\theta(x_{1:T}\vert z)\right]}_\text{Reconstruction Loss}-\underbrace{\vphantom{\mathop\mathbb{E}_{z\sim~q_\phi(\cdot\vert x_{1:T})}\left[\log p_\theta(x_{1:T}\vert z)\right]}\kld(q_\phi(z\vert x_{1:T})\Vert p_\theta(z))}_\text{KL divergence}
\end{aligned}
\label{eq:vae_elbo}
\end{equation}

\noindent where we have introduced the variational posterior $q$ parametrized by new parameters $\phi$. $q$ is referred to as an ``inference model'' as it encodes an input into the latent space. We also have the generative model probabilities $p$ parametrized by $\theta$. The parameters are trained in a way that reflects a classical autoencoder framework: encode the input into a latent space, decode the latent space to reconstruct the input. These models are therefore referred to as ``variational autoencoders''.

The lower bound consists of two terms: reconstruction loss and KL divergence.
The KL divergence term provides a regularizing effect during learning by ensuring that the learned posterior remains close to the prior over the latent variables.

\subsection{Variational Sequential Labelers}

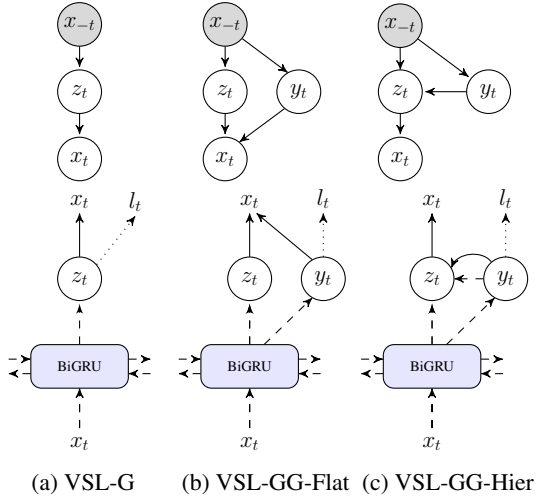
\begin{figure}[t]
    \centering
    \begin{subfigure}[b]{.15\textwidth}
    \centering
        \begin{tikzpicture}
  [node distance=1.5em,
  start chain=going above,
  scale=0.5,
  every node/.style={scale=0.5}]
	 \node[bignode, gen_circle, fill=black!15] (x1) {$x_{-t}$};
	 \node[bignode, gen_circle, below=0.8em of x1, join] (z) {$z_t$};
	 \node[bignode, gen_circle, below=0.8em of z, join] (x) {$x_t$};
\end{tikzpicture}
    \end{subfigure}%
    \begin{subfigure}[b]{.15\textwidth}
    \centering
        \begin{tikzpicture}
  [node distance=1.5em,
  start chain=going above,
  scale=0.5,
  every node/.style={scale=0.5}]
	 \node[bignode, gen_circle, fill=black!15] (x1) {$x_{-t}$};
	 \node[bignode, gen_circle, below=0.8em of x1, join] (z) {$z_t$};
	 \node[bignode, gen_circle, below=0.8em of z, join] (x) {$x_t$};
	 \node[bignode, gen_circle, right=1em of z] (y) {$y_t$};
	 
     \draw[->,thin] (y.230) -> (x.50);
     \draw[->,thin] (x1.320) -> (y.120);
\end{tikzpicture}
    \end{subfigure}%
    \begin{subfigure}[b]{.15\textwidth}
    \centering
        \begin{tikzpicture}
  [node distance=1.5em,
  start chain=going above,
  scale=0.5,
  every node/.style={scale=0.5}]
	 \node[bignode, gen_circle, fill=black!15] (x1) {$x_{-t}$};
	 \node[bignode, gen_circle, below right=1.2em and 1.95em of x1, join] (y) {$y_t$};
	 \node[bignode, gen_circle, left=1.5em of y, join] (z) {$z_t$};
	 \node[bignode, gen_circle, below=0.8em of z, join] (x) {$x_t$};
     \draw[->,thin] (x1.south) -> (z.north);
\end{tikzpicture}
    \end{subfigure}
    \begin{subfigure}[b]{.15\textwidth}
    \centering
        \begin{tikzpicture}
  [node distance=1.5em,
  start chain=going above,
  scale=0.5,
  every node/.style={scale=0.5}]
 \node[punktchain] (l1) {BiGRU};
 \node[bignode, infer_circle, on chain] (c2) {$z_t$};
 \node[bignode, below=1.5em of l1] (in) {$x_t$};
 \node[bignode, above=1.5em of c2] (out) {$x_t$};
 \node[bignode, right=1.0em of out] (out2) {$l_t$};
 \draw[->,thin] (c2.north) -> (out.south);
 \draw[->,thin,dotted] (c2.400) -> (out2.south);
 \draw[->,thin, dashed] (l1.south) |-+(0,-3em)-> (l1);
 \draw[->,thin, dashed] ($(l1.west)+(-1.5em,0.5em)$) -> ($(l1.west)+(0.0em,0.5em)$);
  \draw[->,thin, dashed] ($(l1.east)+(0.0em,0.5em)$) -> ($(l1.east)+(1.5em,0.5em)$);
 \draw[->,thin, dashed]  ($(l1.west)+(0.0em,-0.5em)$) -> ($(l1.west)+(-1.5em,-0.5em)$);
  \draw[->,thin, dashed] ($(l1.east)+(1.5em,-0.5em)$) -> ($(l1.east)+(0.0em,-0.5em)$);
\end{tikzpicture}
        \caption{\vsmg}
        \label{model:gauss}
    \end{subfigure}%
    \begin{subfigure}[b]{.15\textwidth}
    \centering
        \begin{tikzpicture}
  [node distance=1.5em,
  start chain=going above,
  scale=0.5,
  every node/.style={scale=0.5}]
 \node[punktchain] (l1) {BiGRU};
 \node[bignode, infer_circle, on chain] (c2) {$z_t$};
 \node[bignode, infer_circle, right=1em of c2] (c3) {$y_t$};
 \node[bignode, below=1.5em of l1] (in) {$x_t$};
 \node[bignode, above=1.5em of c2] (out) {$x_t$};
 \node[bignode, above=1.5em of c3] (out2) {$l_t$};
 \draw[->,thin] (c2.north) -> (out.south);
 \draw[->,thin,dotted] (c3.north) -> (out2.south);
 \draw[->,thin, dashed] ($(l1.north)+(1.0em,0)$) -> (c3.240);
 \draw[->,thin] (c3.120) -> (out.300);
 \draw[->,thin, dashed] (l1.south) |-+(0,-3em)-> (l1);
 \draw[->,thin, dashed] ($(l1.west)+(-1.5em,0.5em)$) -> ($(l1.west)+(0.0em,0.5em)$);
  \draw[->,thin, dashed] ($(l1.east)+(0.0em,0.5em)$) -> ($(l1.east)+(1.5em,0.5em)$);
 \draw[->,thin, dashed]  ($(l1.west)+(0.0em,-0.5em)$) -> ($(l1.west)+(-1.5em,-0.5em)$);
  \draw[->,thin, dashed] ($(l1.east)+(1.5em,-0.5em)$) -> ($(l1.east)+(0.0em,-0.5em)$);
\end{tikzpicture}
        \caption{\vsmggflat}
        \label{model:2gauss_ind}
    \end{subfigure}%
    \begin{subfigure}[b]{.15\textwidth}
    \centering
        \begin{tikzpicture}
  [node distance=1.5em,
  start chain=going above,
  scale=0.5,
  every node/.style={scale=0.5}]
 \node[punktchain] (l1) {BiGRU};
 \node[bignode, infer_circle, on chain] (c2) {$z_t$};
 \node[bignode, infer_circle, right=1em of c2] (c3) {$y_t$};
 \node[bignode, below=1.5em of l1] (in) {$x_t$};
 \node[bignode, above=1.5em of c2] (out) {$x_t$};
 \node[bignode, above=1.5em of c3] (out2) {$l_t$};
 \draw[->,thin] (c2.north) -> (out.south);
 \draw[->,thin,dotted] (c3.north) -> (out2.south);
 \draw[->,thin, dashed] (c3.west) -> (c2.east);
 \draw[->,thin] (c3.130) to [bend right=60] (c2.30);
 \draw[->,thin, dashed] ($(l1.north)+(1.0em,0)$) -> (c3.240);
 \draw[->,thin, dashed] (l1.south) |-+(0,-3em)-> (l1);
 \draw[->,thin, dashed] ($(l1.west)+(-1.5em,0.5em)$) -> ($(l1.west)+(0.0em,0.5em)$);
  \draw[->,thin, dashed] ($(l1.east)+(0.0em,0.5em)$) -> ($(l1.east)+(1.5em,0.5em)$);
 \draw[->,thin, dashed]  ($(l1.west)+(0.0em,-0.5em)$) -> ($(l1.west)+(-1.5em,-0.5em)$);
  \draw[->,thin, dashed] ($(l1.east)+(1.5em,-0.5em)$) -> ($(l1.east)+(0.0em,-0.5em)$);
\end{tikzpicture}
        \caption{\vsmgghier}
        \label{model:2gauss_cond}
    \end{subfigure}
    \caption{Variational sequential labelers. The first row shows the original graphical models of each variant 
    where shaded circles are observed variables. The second row shows how we perform inference and learning, showing inference models (in dashed lines), generative models (in solid lines), and classifier (in dotted lines). All models are trained to maximize $p_\theta (x_t\vert x_{-t})$ and predict the label $l_t$.}
    \label{fig1}
\end{figure}

We now introduce variational sequential labelers (VSLs) and propose several variants for sequence labeling tasks. Although the latent structure varies, a VSL maximizes the conditional probability of $p_\theta(x_t\vert x_{-t})$ and minimizes a classification loss using the latent variables as the input to the classifier. Unlike VAEs, VSLs do not autoencode the input, so they are more similar
to recent conditional variational formulations~\citep{sohn2015learning,miao2016neural,zhou2017multi}.
Intuitively, the VSL variational objective is to find the information that is useful for predicting the word $x_t$ from its surrounding context, which has similarities to objectives for learning word embeddings~\citep{Collobert:2011:NLP:1953048.2078186,mikolov2013efficient}. This objective serves as regularization for the labeled data and as an unsupervised objective for the unlabeled data.

All of our models use latent variables for each position in the sequence. These characteristics are shown in the visual depictions of our models in Figure~\ref{fig1}.
We consider variants with multiple latent variables per time step and attach the classifier to only particular variables. This causes the different latent variables to capture different characteristics.

In the following sections, we will describe various latent variable configurations that we will evaluate empirically in subsequent sections.

\subsection{Single Latent Variable}

We begin by defining a basic VSL and corresponding parametrization, which will also be used in other variants. This first model (which we call \vsmg and show in Figure~\ref{model:gauss}) has a Gaussian latent variable at each time step. \vsmg uses two training objectives; the first is similar to the lower bound on log-likelihood used by VAEs:

\begin{equation}
\begin{aligned}
    &\log~p_\theta(x_t\vert x_{-t})\geq\mathop\mathbb{E}_{z_t\sim q_\phi(\cdot\vert~x_{1:T},t)}[\log p_\theta(x_t\vert~z_t)-\\
    &\log\frac{q_\phi(z_t\vert~x_{1:T},t)}{p_\theta(z_t\vert~x_{-t})} ]=\mathop\mathbb{E}_{z_t\sim q_\phi(\cdot\vert~x_{1:T},t)}[\log p_\theta(x_t\vert~z_t)]\\
    &-\kld(q_\phi(z_t\vert~x_{1:T},t)\Vert~p_\theta(z_t\vert~x_{-t}))=U_0(x_{1:T}, t)
\end{aligned}
\label{eq:rec_gaussian}
\end{equation}

\noindent \vsmg additionally uses a classifier $f$ on the latent variable $z_t$ which is trained with the following objective:
\begin{equation}
\begin{aligned}
C_0(x_{1:T}, l_{t}) = \mathop\mathbb{E}_{z_t\sim q_\phi(\cdot\vert x_{1:T},t)}
[-\log f(l_t\vert z_t)]
\end{aligned}
\label{eq:ce_rec}
\end{equation}

\noindent The final loss is
\begin{equation}
L(x_{1:T},l_{1:T}) = \sum_{t=1}^T[C_0(x_{1:T}, l_{t}) - \alpha U_0(x_{1:T}, t)]\nonumber
\end{equation}
where $\alpha$ is a trade-off hyperparameter. $\alpha$ is set to zero during supervised training but it is tuned based on development set performance during semi-supervised training. The same procedure is adopted for the other VSL models below.

For the generative model, we parametrize $p_\theta(x_t\vert z_t)$ as a feedforward neural network with two hidden layers and ReLU~\cite{nair2010rectified}
as activation function. As reconstruction loss, we use cross-entropy over the words in the vocabulary.
We defer the descriptions of the parametrization of $p_\theta(z_t\vert~x_{-t})$ to Section~\ref{prior}.

We now discuss how we parametrize the inference model $q_{\phi}(z_t\vert x_{1:T},t)$.
We use a bidirectional gated recurrent unit (BiGRU; \citealp{chung2014empirical}) network to produce a  hidden vector $h_t$ at position $t$. The BiGRU is run over the input $x_{1:T}$, where each $x_t$ is the concatenation of a word embedding and the concatenated final hidden states from a character-level BiGRU.
The inference model $q_{\phi}(z_t\vert x_{1:T},t)$ is then a single layer feedforward neural network that uses $h_t$ as input.
When parametrizing the posterior over latent variables in the following models below, we use this same procedure to produce hidden vectors with a BiGRU and then use them as input to feedforward networks. The structure of our inference model is similar to those used in previous state-of-the-art models for sequence labeling~\cite{lample2016neural,yang2017transfer}.

In order to focus more on the effect of our variational objective, the classifier we use is always the same as our baseline model (see Section \ref{baseline}), which is a one layer feedforward neural network without a hidden layer, and it is also used in test-time prediction.

\subsection{Flat Latent Variables}
We next consider ways of factorizing the functionality of the latent variable into label-specific and other word-specific information.
We introduce \vsmggflat (shown in Figure~\ref{model:2gauss_ind}), which has two conditionally independent Gaussian latent variables at each time step, denote $z_t$ and $y_t$ for time step $t$.

The variational lower bound is derived as follows:
\begin{equation}
\begin{aligned}
    \log~&p_\theta(x_t\vert x_{-t})\geq\\
    &\mathop\mathbb{E}_{z_t,y_t\sim q_\phi(\cdot\vert x_{1:T},t)}[\log p_\theta(x_t\vert~z_t,y_t) \\&-\log\frac{q_\phi(z_t\vert x_{1:T},t)}{p_\theta(z_t\vert x_{-t})}
    -\log\frac{q_\phi(y_t\vert x_{1:T},t)}{p_\theta(y_t\vert x_{-t})}]\\
    &=\mathop\mathbb{E}_{z_t,y_t\sim q_\phi(\cdot\vert x_{1:T},t)}[\log p_\theta(x_t\vert z_t,y_t)]\\
    &-\kld(q_\phi(z_t\vert x_{1:T},t)\Vert p_\theta(z_t\vert x_{-t}))\\
    &-\kld(q_\phi(y_t\vert x_{1:T},t)\Vert p_\theta(y_t\vert x_{-t}))
    \\&=U_1(x_{1:T}, t)
\end{aligned}
\label{eq:elbo1}
\end{equation}

\noindent The classifier $f$ is on the latent variable $y_t$ and its loss is
\begin{equation}
\begin{aligned}
C_1(x_{1:T}, l_{t}) = \mathop\mathbb{E}_{y_t\sim q_\phi(\cdot\vert x_{1:T},t)}
[-\log f(l_t\vert y_t)]
\end{aligned}
\label{eq:ce1}
\end{equation}

\noindent The final loss for the model is
\begin{equation}
L(x_{1:T},l_{1:T}) = \sum_{t=1}^T[C_1(x_{1:T}, l_{t}) - \alpha U_1(x_{1:T},t)]
\end{equation}
Where $\alpha$ is a trade-off hyperparameter.

Similarly to the VSL-G model, $q_\phi(z_t\vert x_{1:T},t)$ and $q_\phi(y_t\vert x_{1:T},t)$ are parametrized by single layer feedforward neural networks using the hidden state $h_t$ as input.

\subsection{Hierarchical Latent Variables}

We also explore hierarchical relationships among the latent variables.
In particular, we introduce the \vsmgghier model which has two Gaussian latent variables with the  hierarchical structure shown in Figure~\ref{model:2gauss_cond}. This model encodes the intuition that the word-specific latent information $z_t$ may differ depending on the class-specific information of $y_t$.

For this model, the derivations are similar to Equations (\ref{eq:elbo1}) and (\ref{eq:ce1}). The first is:
\begin{equation}
\begin{aligned}
    \log~&p_\theta(x_t\vert x_{-t})\geq\\
    &\mathop\mathbb{E}_{z_t,y_t\sim q_\phi(\cdot\vert x_{1:T},t)}[\log p_\theta(x_t\vert z_t) \\&-\log\frac{q_\phi(z_t\vert y_t,x_{1:T},t)}{p_\theta(z_t\vert y_t,x_{-t})}
    -\log\frac{q_\phi(y_t\vert x_{1:T},t)}{p_\theta(y_t\vert x_{-t})} ]\\
    &=\mathop\mathbb{E}_{z_t,y_t\sim q_\phi(\cdot\vert x_{1:T},t)}[\log p_\theta(x_t\vert z_t)]\\
    &-KL(q_\phi(z_t\vert y_t,x_{1:T},t)\Vert p_\theta(z_t\vert y_t,x_{-t}))\\
    &-KL(q_\phi(y_t\vert x_{1:T},t)\Vert p_\theta(y_t\vert x_{-t}))\\
    &=U_2(x_{1:T},t)
\end{aligned}
\label{eq:elbo2}
\end{equation}

\noindent The classifier $f$ uses $y_t$ as input and is trained with the following loss:
\begin{equation}
\begin{aligned}
C_2(x_{1:T}, l_{t}) = \mathop\mathbb{E}_{y_t\sim q_\phi(\cdot\vert x_{1:T},t)}
[-\log f(l_t\vert y_t)]
\end{aligned}
\label{eq:ce2}
\end{equation}

\noindent Note that $C_1$ and $C_2$ have the same form.
The final loss is
\begin{equation}
\begin{aligned}
L(x_{1:T},l_{1:T}) &= \sum_{t=1}^T[ C_2(x_{1:T}, l_{t}) - \alpha~U_2(x_{1:T}, t)]
\end{aligned}
\end{equation}
Where $\alpha$ is a trade-off hyperparameter.

The hierarchical posterior $q_{\phi}(z_t |y_t,x_{1:T},t)$ is parametrized by concatenating the hidden vector $h_t$ and the random variable $y_t$ and then using them as input to a single layer feedforward 
network.

\subsection{Parametrization of Priors}
\label{prior}

Traditional variational models assume extremely simple priors (e.g., multivariate standard Gaussian distributions). Recently there have been efforts to learn the prior and posterior jointly during training~\cite{fraccaro2016sequential,serban2017piecewise,tomczak2017vae}.
In this paper, we follow this same idea  but we do not explicitly parametrize the prior  $p_\theta(z_t\vert x_{-t})$. This is partially due to the lack of computationally-efficient parametrization options for $p_\theta(z_t\vert x_{-t})$. In addition, since we are not seeking to do generation with our learned models, we can let part of the generative model be parametrized implicitly.

More specifically, the approach we use is to learn the priors by updating them iteratively. During training, we first initialize the priors of all examples as multivariate standard Gaussian distributions. As  training proceeds, we use the last optimized posterior as our current prior based on a particular ``update frequency'' (see supplementary material for more details).

Our learned priors are implicitly modeled as
\begin{equation}
\begin{aligned}
   &p^k_\theta(z_t\vert x_{-t}) \approx \\
   &\sum_{x}~q^{k-1}_\phi(z_t\vert X_t = x,x_{-t}, t)p_\text{data}(X_t = x \vert x_{-t})
\end{aligned}
\label{eqn:prior}
\end{equation}
\noindent

where $p_\text{data}$ is the empirical data distribution, $X_t$ is a random variable corresponding to the observation at position $t$, and $k$ is the prior update time step.
The intuition here is that the prior is obtained by marginalizing over values for the missing observation represented by the random variable $X_t$. The posterior $q^{k-1}_\phi$ is as defined in our latent variable models.
We assume $p_\text{data}(X_t=x\vert x_{-t})=0$ for $x_{1:T}\notin\text{training set}$.
For context $x_{-t}$ that can pair with multiple values of $X_t$, its prior is the data-dependent weighted average posterior.
For simplicity of implementation and efficient computation, however, if context $x_{-t}$ can pair with multiple values in our training data, we ignore this fact and simply use instance-dependent posteriors.
Another way to view this is as conditioning on the index of the training examples while parametrizing the above. That is

\begin{equation}
   p^{k,i}_\theta(z_t\vert x_{-t})\leftarrow q^{k-1,i}_\phi(z_t\vert x_{1:T}, t)\label{eqn:sim-prior}
\end{equation}

\noindent where $i$ is the index of the instance.

\subsection{Training}
In this subsection, we introduce techniques we have used to address difficulties during training.

\paragraph{Reparametrization Trick.}
It is challenging to use gradient descent for a random variable as it involves a non-differentiable sampling procedure.~\citet{kingma2013auto} introduced a reparametrization trick to tackle this problem.
They parametrize a Gaussian random variable $z$ as $u_\varphi(x) + g_\psi(x)\circ\epsilon$ where
$\epsilon\sim\mathcal{N}(0,\mathbf{I})$
\qtcomments{$\mathcal{N}(0,\mathbf{I})$ seems more precise.} and $u_\varphi(x)$, $g_\psi(x)$ are deterministic and differentiable functions,
so the gradient can go through
$u_\varphi(\cdot)$ and $g_\psi(\cdot)$.
In our experiments, we use one sample for each time step during training.  For evaluation at test time, we use the mean value $u_\varphi(x)$.

\paragraph{KL Divergence Weight Annealing.} Although the use of prior updating

lets us avoid tuning the weight of the KL divergence, the simple priors can still hinder learning during the initial stages of training. To address this, we follow the method described by~\citet{bowman2016generating} to add weights to all KL divergence terms and anneal the weights from a small value to 1.

\section{Experiments}

We describe key details of our experimental setup in the subsections below but defer details about hyperparameter tuning to the supplementary material. Our implementation is available at \url{https://github.com/mingdachen/vsl}

\subsection{Datasets}
We evaluate our model on the CoNLL 2003 English NER dataset~\cite{tjong2003introduction} and 7 POS tagging datasets: the Twitter tagging dataset of~\citet{gimpel-11a} and~\citet{owoputi-13}, and 6 languages from the Universal Dependencies (UD) 1.4 dataset~\cite{mcdonald2013universal}.

\paragraph{Twitter POS Dataset.} The Twitter dataset has 25 tags. We use \textsc{Oct27Train} and \textsc{Oct27Dev} as the training set, \textsc{Oct27Test} as the development set, and \textsc{Daily547} as the test set. We randomly sample \{1k, 2k, 3k, 4k, 5k, 10k, 20k, 30k, 60k\} tweets from 56 million English tweets as our unlabeled data and tune the amount of unlabeled data based on development set accuracy.
\paragraph{UD POS Datasets.} The UD datasets have 17 tags. We use French, German, Spanish, Russian, Indonesian and Croatian. We follow the same setup as~\citet{zhang2017semi}, randomly sampling 20\% of the original training set as our labeled data and 50\% as unlabeled data. There is no overlap between the labeled and unlabeled data. See~\citet{zhang2017semi} for more details about the setup.
\paragraph{NER Dataset.} We use the BIOES labeling scheme and report micro-averaged $F_{1}$.
We preprocessed the text by replacing all digits with 0. We randomly sample 10\% of the original training set as our labeled data and 50\% as unlabeled data. We also ensure there is no overlap between the labeled and unlabeled data.

\subsection{Pretrained Word Embeddings}
For all experiments, we use pretrained 100-dimensional word embeddings. For Twitter, we trained skip-gram embeddings~\cite{mikolov2013efficient} on a dataset of 56 million English tweets.
For the UD datasets, we trained skip-gram embeddings on Wikipedia for each of the six languages. For NER, we use 100-dimensional pretrained GloVe~\cite{pennington2014glove} embeddings. Our models perform better with word embeddings kept fixed during training while for the baselines the word embeddings are fine tuned as this improves the baseline performance.\klcomment{added "as this..."}

\subsection{Baselines}
\label{baseline}
Our primary baseline is a BiGRU tagger where the input consists of the concatenation of a word embedding and the concatenation of the final hidden states of a character-level BiGRU. This BiGRU architecture is identical to that used in the inference networks in our VSL models.
Predictions are made based on a linear transformation given the current hidden state. The output dimensionality of the transformation is task-dependent (e.g., 25 for Twitter tagging). We use the standard per-position cross entropy loss for training.

We also report results from the best systems from \citet{zhang2017semi}, namely the NCRF and NCRF-AE models. Both use feedforward networks as encoders and conditional random field layers for capturing sequential information. The NCRF-AE model additionally can benefit from unlabeled data.

\section{Results}

\begin{table}[t]
\setlength{\tabcolsep}{3pt}
\begin{subtable}{.5\textwidth}
\centering
\begin{tabular}{l|c|c|c|c}
        & \multicolumn{2}{c|}{dev.} & \multicolumn{2}{c}{test} \\
        & \multicolumn{1}{c}{acc.} & UL$\Delta$ & \multicolumn{1}{c}{acc.} & UL$\Delta$ \\
\hline
BiGRU baseline     & 90.8  & -    & 90.6 & -  \\
\vsmg              & 91.1  &  +0.1 &   -  & -  \\
\vsmggflat            & 91.4  &  +0.1 &   -  & -  \\
\vsmgghier            & \textit{\textbf{91.6}}  & \textit{\textbf{+0.3}}  &   \textit{\textbf{91.6}}   &  \textit{+0.3} \\
\end{tabular}
\caption{Twitter tagging accuracies (\%)}
\label{twitter-res}
\end{subtable}

\begin{subtable}{.5\textwidth}
\centering
\begin{tabular}{l|c|c|c|c}
        & \multicolumn{2}{c|}{dev.} & \multicolumn{2}{c}{test} \\
        & \multicolumn{1}{c}{$F_{1}$} & UL$\Delta$ & \multicolumn{1}{c}{$F_{1}$} & UL$\Delta$ \\
\hline
BiGRU baseline     & 87.6  & -    & 83.7 & -  \\
\vsmg              & 87.8  & +0.1  &   -  & -  \\
\vsmggflat         & 88.0  & +0.1  &   -  & -  \\
\vsmgghier         & \textit{\textbf{88.4}}  & \textit{\textbf{+0.2}} &  \textit{\textbf{84.7}}    &    \textit{+0.0} \\
\end{tabular}
\caption{NER $F_{1}$ score (\%)}
\label{ner-res}
\end{subtable}
\caption{For dev and test, we show results when only using labeled data and the change in performances (``UL$\Delta$'') when adding unlabeled data. Bold is highest in each column. Italic is the best model including unlabeled data. We only show test results for the baseline and our best-performing model, which achieves 91.9\% accuracy on the Twitter test set and 84.7\% $F_1$ on the NER test set when using unlabeled data.}
\end{table}

Table~\ref{twitter-res} shows results on the Twitter development and test sets.
All of our VSL models outperform the baseline and our best VSL models outperform the BiGRU baseline by 0.8\---1\% absolute. When comparing different latent variable configurations, we find that a hierarchical structure performs best. Without unlabeled data, our models already outperform the BiGRU baseline.
Adding unlabeled data enlarges the gap between the baseline and our models by up to 0.1\---0.3\% absolute.

\begin{table*}[t]
\setlength{\tabcolsep}{4pt}
\centering
\begin{tabular}{l|c|c|c|c|c|c|c|c|c|c|c|c}
        & \multicolumn{2}{c|}{French} & \multicolumn{2}{c|}{German} & \multicolumn{2}{c|}{Indonesian} & \multicolumn{2}{c|}{Spanish} & \multicolumn{2}{c|}{Russian} & \multicolumn{2}{c}{Croatian} \\
        & \multicolumn{1}{c}{acc.} & UL$\Delta$
        & \multicolumn{1}{c}{acc.} & UL$\Delta$
        & \multicolumn{1}{c}{acc.} & UL$\Delta$
        & \multicolumn{1}{c}{acc.} & UL$\Delta$
        & \multicolumn{1}{c}{acc.} & UL$\Delta$
        & \multicolumn{1}{c}{acc.} & UL$\Delta$ \\
\hline
NCRF               & 93.4 &   -  & 90.4 &   -   & 88.4 &   -  & 91.2 &   -  & 86.6 &   -  & 86.1 &  -   \\
NCRF-AE            & 93.7 & +0.2 & 90.8 & +0.2  & 89.1 & +0.3 & 91.7 & +0.5 & 87.8 & +1.1 & 87.9 & +1.2 \\
\hline
BiGRU baseline     & 95.9 &   -   & 92.6 &   -  & 92.2 &   -   & 94.7 &   -  & 95.2 &   -  & 95.6 &  -   \\
\vsmg              & 96.1 & +0.0  & 92.8  & +0.0 & 92.3 & +0.0 & 94.8 & +0.1 & 95.3 & +0.0 & 95.6 &  +0.1  \\
\vsmggflat         & 96.1 & +0.0  & 93.0  & \textbf{+0.1} & 92.4 & \textbf{+0.1} & 95.0 & +0.1 & 95.5 & \textbf{+0.1} & 95.8 & +0.1 \\
\vsmgghier         & \textit{\textbf{96.4}} & \textit{\textbf{+0.1}}  & \textit{\textbf{93.3}} & \textit{\textbf{+0.1}} & \textit{\textbf{92.8}} & \textit{\textbf{+0.1}} & \textit{\textbf{95.3}} & \textit{\textbf{+0.2}} & \textit{\textbf{95.9}} & \textit{\textbf{+0.1}} & \textit{\textbf{96.3}} & \textit{\textbf{+0.2}} \\
\end{tabular}
\caption{Tagging accuracies (\%) on UD test sets. For each language, we show test accuracy (``acc.'') when only using labeled data and the change in test accuracy (``UL$\Delta$'') when adding unlabeled data. Results for NCRF and NCRF-AE are from \citet{zhang2017semi}, though results are not strictly comparable because we used pretrained word embeddings for all languages on Wikipedia. Bold is highest in each column, excluding the NCRF variants. Italic is the best accuracy including the unlabeled data.
}
\label{ud-res}
\end{table*}

Table~\ref{ner-res} shows results on the CoNLL 2003 NER development and test sets. We observe similar trends as in the Twitter data, except that the model does not show improvement on the test set when adding unlabeled data.

Table~\ref{ud-res} shows our results on the UD datasets. The trends are broadly consistent with those of Table~\ref{twitter-res} and~\ref{ner-res}. The best performing models use hierarchical structure in the latent variables.
There are some differences across languages. For French, German, Indonesian and Russian, \vsmg does not show improvement when using unlabeled data. This may be resolved with better tuning, since the model actually shows improvement on the dev set.

Note that results reported by~\citet{zhang2017semi} and ours are not strictly comparable as their word embeddings were only pretrained on the UD training sets while ours were pretrained on Wikipedia. Nonetheless, they also mentioned that using embeddings pretrained on larger unlabeled data did not help. We include these results to show that our baselines are indeed strong compared to prior results reported in the literature.

\section{Discussion}
\qtcomment{The discussion part looks interesting to me. I would suggest to put subsection ~\ref{subsec:discuss-multi-task} behind subsection ~\ref{subsec:discuss-latent-hierarchy}. The logic is single-flat-hierarchy-different positions in hierarchy}

\subsection{Effect of Position of Classification Loss
\label{subsec:discuss-multi-task}}

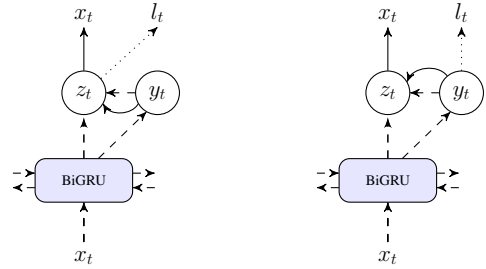
\begin{figure}[t]
    \centering
    \begin{subfigure}[t]{.25\textwidth}
    \centering
        \begin{tikzpicture}
  [node distance=1.5em,
  start chain=going above,
  scale=0.5,
  every node/.style={scale=0.5}]
 \node[punktchain] (l1) {BiGRU};
 \node[bignode, infer_circle, on chain] (c2) {$z_t$};
 \node[bignode, infer_circle, right=1em of c2] (c3) {$y_t$};
 \node[bignode, below=1.5em of l1] (in) {$x_t$};
 \node[bignode, above=1.5em of c2] (out) {$x_t$};
 \node[bignode, above=1.5em of c3] (out2) {$l_t$};
 \draw[->,thin] (c2.north) -> (out.south);
 \draw[->,thin,dotted] (c2.400) -> (out2.south);
 \draw[->,thin, dashed] (c3.west) -> (c2.east);
 \draw[->,thin] (c3.210) to [bend left=60] (c2.330);
 \draw[->,thin, dashed] ($(l1.north)+(1.0em,0)$) -> (c3.240);
 \draw[->,thin, dashed] (l1.south) |-+(0,-3em)-> (l1);
 \draw[->,thin, dashed] ($(l1.west)+(-1.5em,0.5em)$) -> ($(l1.west)+(0.0em,0.5em)$);
  \draw[->,thin, dashed] ($(l1.east)+(0.0em,0.5em)$) -> ($(l1.east)+(1.5em,0.5em)$);
 \draw[->,thin, dashed]  ($(l1.west)+(0.0em,-0.5em)$) -> ($(l1.west)+(-1.5em,-0.5em)$);
  \draw[->,thin, dashed] ($(l1.east)+(1.5em,-0.5em)$) -> ($(l1.east)+(0.0em,-0.5em)$);
\end{tikzpicture}
        \caption{\vsmgghier with classification loss on $z$}
        \label{fig:condz}
    \end{subfigure}%
    \begin{subfigure}[t]{.25\textwidth}
    \centering
        \begin{tikzpicture}
  [node distance=1.5em,
  start chain=going above,
  scale=0.5,
  every node/.style={scale=0.5}]
 \node[punktchain] (l1) {BiGRU};
 \node[bignode, infer_circle, on chain] (c2) {$z_t$};
 \node[bignode, infer_circle, right=1em of c2] (c3) {$y_t$};
 \node[bignode, below=1.5em of l1] (in) {$x_t$};
 \node[bignode, above=1.5em of c2] (out) {$x_t$};
 \node[bignode, above=1.5em of c3] (out2) {$l_t$};
 \draw[->,thin] (c2.north) -> (out.south);
 \draw[->,thin,dotted] (c3.north) -> (out2.south);
 \draw[->,thin, dashed] (c3.west) -> (c2.east);
 \draw[->,thin] (c3.130) to [bend right=60] (c2.30);
 \draw[->,thin, dashed] ($(l1.north)+(1.0em,0)$) -> (c3.240);
 \draw[->,thin, dashed] (l1.south) |-+(0,-3em)-> (l1);
 \draw[->,thin, dashed] ($(l1.west)+(-1.5em,0.5em)$) -> ($(l1.west)+(0.0em,0.5em)$);
  \draw[->,thin, dashed] ($(l1.east)+(0.0em,0.5em)$) -> ($(l1.east)+(1.5em,0.5em)$);
 \draw[->,thin, dashed]  ($(l1.west)+(0.0em,-0.5em)$) -> ($(l1.west)+(-1.5em,-0.5em)$);
  \draw[->,thin, dashed] ($(l1.east)+(1.5em,-0.5em)$) -> ($(l1.east)+(0.0em,-0.5em)$);
\end{tikzpicture}
        \caption{\vsmgghier}
        \label{fig:cond}
    \end{subfigure}
    \caption{Comparison of attaching classification loss to different latent variables in \vsmgghier.}
    \label{fig:compare}
\end{figure}

\begin{table}[t]
\setlength{\tabcolsep}{4pt}
\centering
\small
\begin{tabular}{l|c|c|c|c|c|c}
        & \multicolumn{2}{c|}{Twitter} & \multicolumn{2}{c|}{NER} & \multicolumn{2}{c}{UD average} \\
        & \multicolumn{1}{c}{acc.} & UL$\Delta$ & \multicolumn{1}{c}{$F_{1}$} & UL$\Delta$ & \multicolumn{1}{c}{acc.} & UL$\Delta$ \\
\hline

classifier on $y$ & 91.6 & +0.3 & 88.4 & +0.2 & 95.0 & +0.1\\

classifier on $z$ & 91.1 & +0.2 & 87.8 & +0.1 & 94.4 & +0.0 \\

\end{tabular}
\caption{Twitter and NER dev results (\%), UD averaged test accuracies (\%) for two choices of attaching the classification loss to latent variables in the \vsmgghier model. All previous results for \vsmgghier used the classification loss on $y$.}
\label{compare-multitask-loss}
\end{table}

We investigate the effect of attaching the classifier to different latent variables. In particular, for the \vsmgghier model, we compare the attachment of the classifier between $z$ and $y$. See Figure~\ref{fig:compare}. The results in Table~\ref{compare-multitask-loss} suggest that attaching the reconstruction and classification losses to the same latent variable ($z$) harms accuracy although attaching the classifier to $z$ effectively gives the classifier an extra layer.
We can observe why this occurs by looking at the latent variable visualizations in Figure~\ref{fig:tsne-2gdepz}. Compared with Figure~\ref{fig:tsne-2gdep}, where the two variables are more clearly disentangled, the latent variables in Figure \ref{fig:tsne-2gdepz} appear to be capturing highly similar information.

\subsection{Effect of Latent Hierarchy}
\label{subsec:discuss-latent-hierarchy}

\begin{figure}[t]
    \centering
    \begin{subfigure}{.25\textwidth}
        \begin{subfigure}{\textwidth}
        \centering
            \includegraphics[width=.9\linewidth]{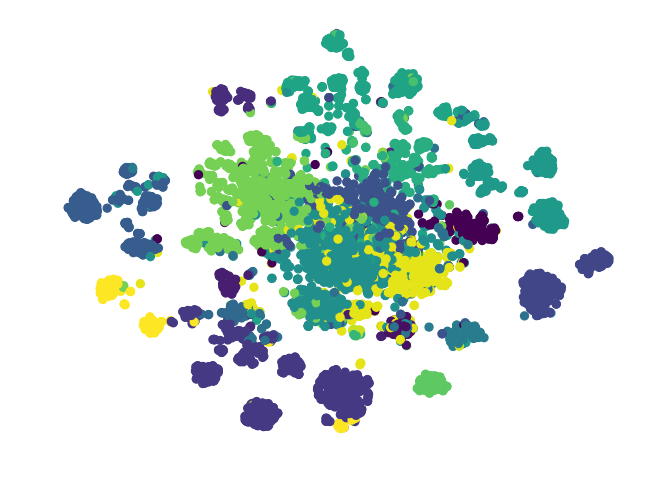}
             \caption{BiGRU Baseline\label{fig:tsne-bigru}}
        \end{subfigure}
        \begin{subfigure}{\textwidth}
        \centering
        \includegraphics[width=.9\linewidth]{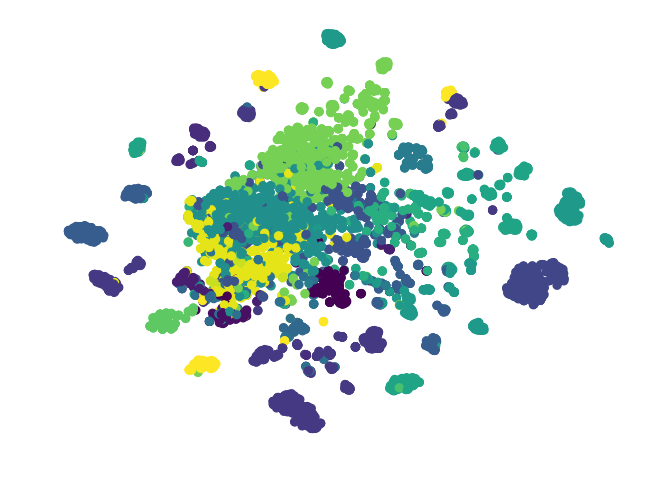}
         \caption{\vsmg\label{fig:tsne-gauss}}
        \end{subfigure}
    \end{subfigure}%
    \begin{subfigure}{.25\textwidth}
    \centering
        \begin{subfigure}{.9\textwidth}
        \includegraphics[width=1.0\linewidth]{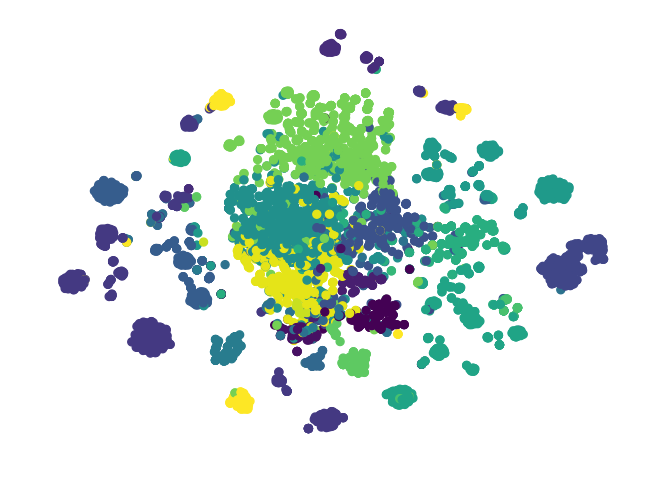}
        \end{subfigure}
        \begin{subfigure}{.9\textwidth}
        \includegraphics[width=1.0\linewidth]{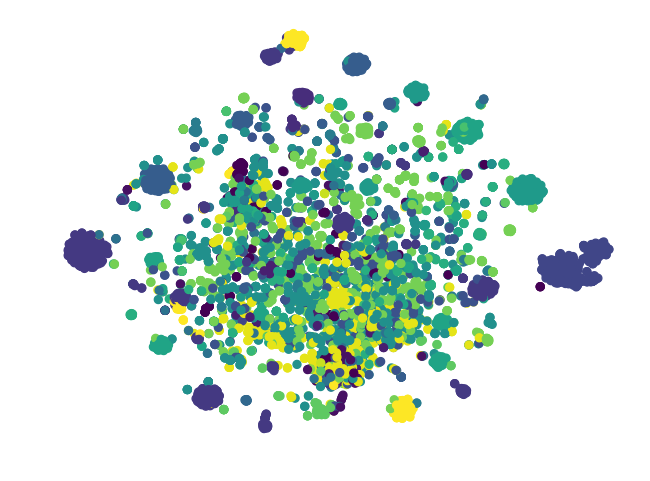}
        \end{subfigure}
     \caption{\vsmggflat \label{fig:tsne-2gind}}
    \end{subfigure}
    \begin{subfigure}{.25\textwidth}
    \centering
        \begin{subfigure}{.9\textwidth}
        \includegraphics[width=1.0\linewidth]{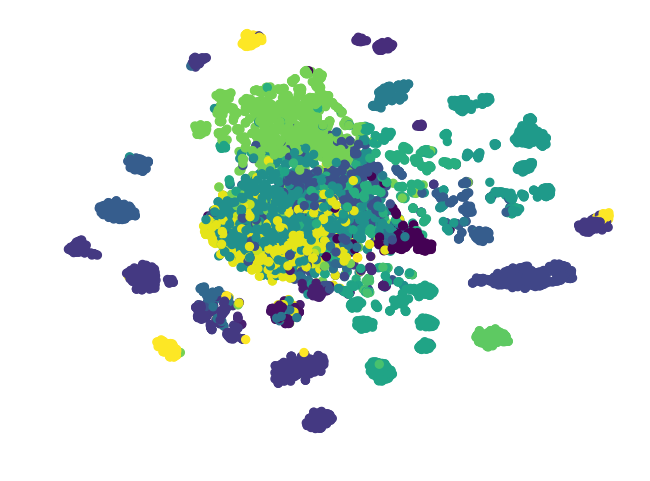}
        \end{subfigure}
        \begin{subfigure}{.9\textwidth}
        \includegraphics[width=1.0\linewidth]{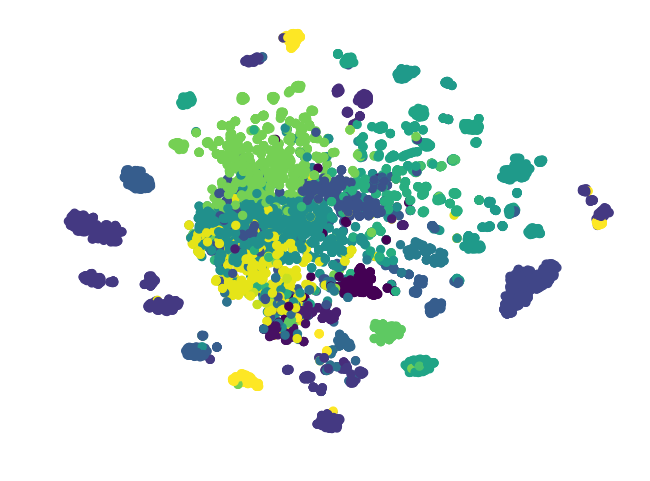}
        \end{subfigure}
     \caption{\vsmgghier, classification loss on $z$  \label{fig:tsne-2gdepz}}
    \end{subfigure}%
    \begin{subfigure}{.25\textwidth}
    \centering
        \begin{subfigure}{.9\textwidth}
        \includegraphics[width=1.0\linewidth]{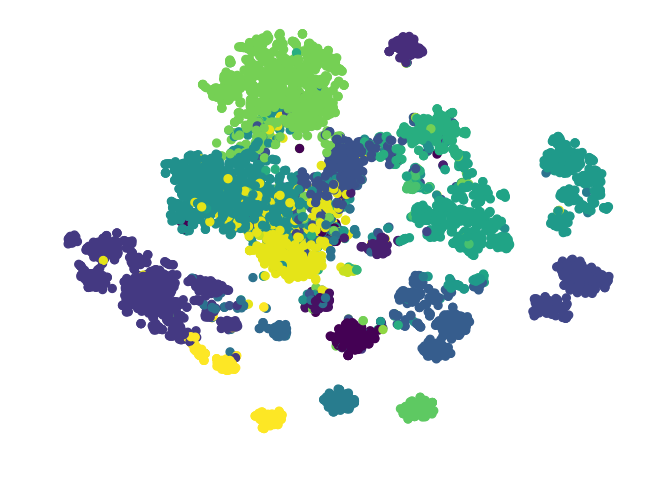}
        \end{subfigure}
        \begin{subfigure}{.9\textwidth}
        \includegraphics[width=1.0\linewidth]{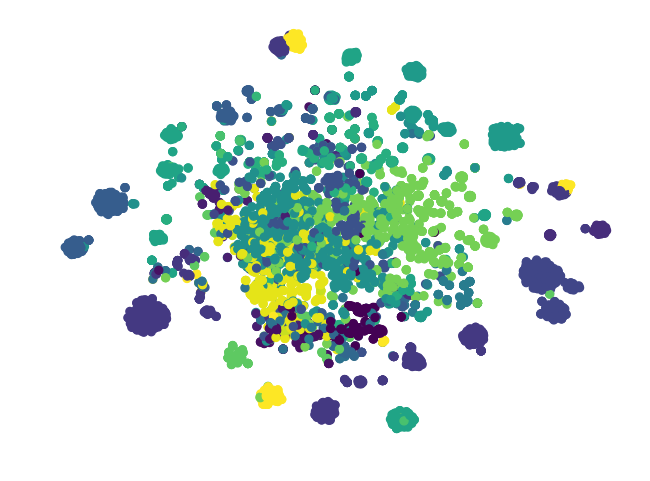}
        \end{subfigure}
     \caption{\vsmgghier \label{fig:tsne-2gdep}}
    \end{subfigure}
\caption{t-SNE visualization of Gaussian latent variables and baseline hidden states for Twitter development set. In plot~\ref{fig:tsne-2gind},~\ref{fig:tsne-2gdepz},  and~\ref{fig:tsne-2gdep}, the upper subplot is latent variable $y$ and the lower is $z$. Each point in the plot is a token and the color represents the true tag of the token.}
\label{fig:gauss_lv}
\end{figure}

To verify our assumption of the latent structure, we visualize the latent space for Gaussian models using t-SNE \cite{maaten2008visualizing} in Figure~\ref{fig:gauss_lv}. The BiGRU baseline (Figure~\ref{fig:tsne-bigru}) and the \vsmg (Figure~\ref{fig:tsne-gauss}) do not show significant differences. However, when using multiple latent variables, the different latent variables capture different characteristics. In the \vsmggflat model (Figure~\ref{fig:tsne-2gind}), the $y$ variable (the upper plot) reflects the clustering of the tagging space much more closely than the $z$ variable (the lower plot). Since both variables are used to reconstruct the word, but only the $y$ variable is trained to predict the tag, it appears that $z$ is capturing other information useful for reconstructing the word.
However, since they are both used for reconstruction, the two spaces show signs of alignment; that is, the ``tag'' latent variable $y$ does not show as clean a separation into tag clusters as the $y$ variable in the \vsmgghier model in Figure~\ref{fig:tsne-2gdep}.

In Figure~\ref{fig:tsne-2gdep} (\vsmgghier), the clustering of words with respect to the tag is clearest.
This may account for the consistently better performance of this model relative to the others. The $z$ variable reflects a space that is conditioned on $y$
but that diverges from it, presumably in order to better reconstruct the word.
The closer the latent variable is to the decoder output, the weaker the tagging information becomes while other word-specific information becomes 
more salient.

Figure~\ref{fig:tsne-2gdepz} shows that \vsmgghier with classification loss on $z$, which consistently underperforms both the \vsmggflat and \vsmgghier models in our experiments, appears to be capturing the same latent space in both variables. Since the $z$ variable is used to both predict the tag and reconstruct the word, it must capture both the tag and word reconstruction spaces, and may be limited by capacity in doing so. The $y$ variable does not seem to be contributing much modeling power, as its space is closely aligned to that of $z$.

\subsection{Effect of Variational Regularization}
\begin{table}[t]
\setlength{\tabcolsep}{4pt}
\centering
\begin{tabular}{l|c|c|c|c}
        & \multicolumn{2}{c|}{Twitter} & \multicolumn{2}{c}{NER} \\
        & \multicolumn{1}{c}{acc.} & no VR & \multicolumn{1}{c}{$F_{1}$} & no VR \\
\hline
BiGRU baseline     & 90.8  &   -  & 87.6 & -   \\
\vsmg              & 91.1  & 90.9 & 87.8 & 87.7 \\
\vsmggflat         & 91.4  & 90.9 & 88.0 & 87.8 \\
\vsmgghier         & 91.6  & 91.0 & 88.4 & 87.9 \\
\end{tabular}
\caption{Results on Twitter and NER dev sets. For each model, we show supervised results for the models with variational regularization (``acc.'' or $F_{1}$) and results when replacing variational components with their deterministic counterparts (``no VR'').}
\label{vr-reg}
\end{table}

We investigate the beneficial effects of variational frameworks (``variational regularization'') by replacing our variational components in VSLs with their deterministic counterparts, which do not have randomness in the latent space and do not use the KL divergence term during optimization.
Note that these BiGRU encoders share the same architectures as their variational posterior counterparts and still use both the classification and
reconstruction losses.
While other subsets of losses could be considered in this comparison, our motivation is to compare two settings that correspond to well-known frameworks. The ``no VR'' setting corresponds roughly to the combination of a classifier and a traditional autoencoder.

We note that these experiments do not use any unlabeled data.

The results in Table~\ref{vr-reg} demonstrate that compared to the baseline BiGRU, adding the reconstruction loss (``\vsmg, no VR'') yields only 0.1 improvement for both Twitter and NER. Although adding hierarchical structure further improves performance, the improvements are small (+0.1 and +0.2 for Twitter and NER respectively).
For \vsmgghier, variational regularization accounts for relatively large differences of 0.6 for Twitter and 0.5 for NER. These results show that the improvements do not come solely from adding a reconstruction objective to the learning procedure. In limited preliminary experiments, we did not find a benefit from adding unlabeled data under the ``no VR'' setting.

\subsection{Effect of Unlabeled Data}

\begin{figure}[t]
    \centering
    \includegraphics[width=1.1\linewidth]{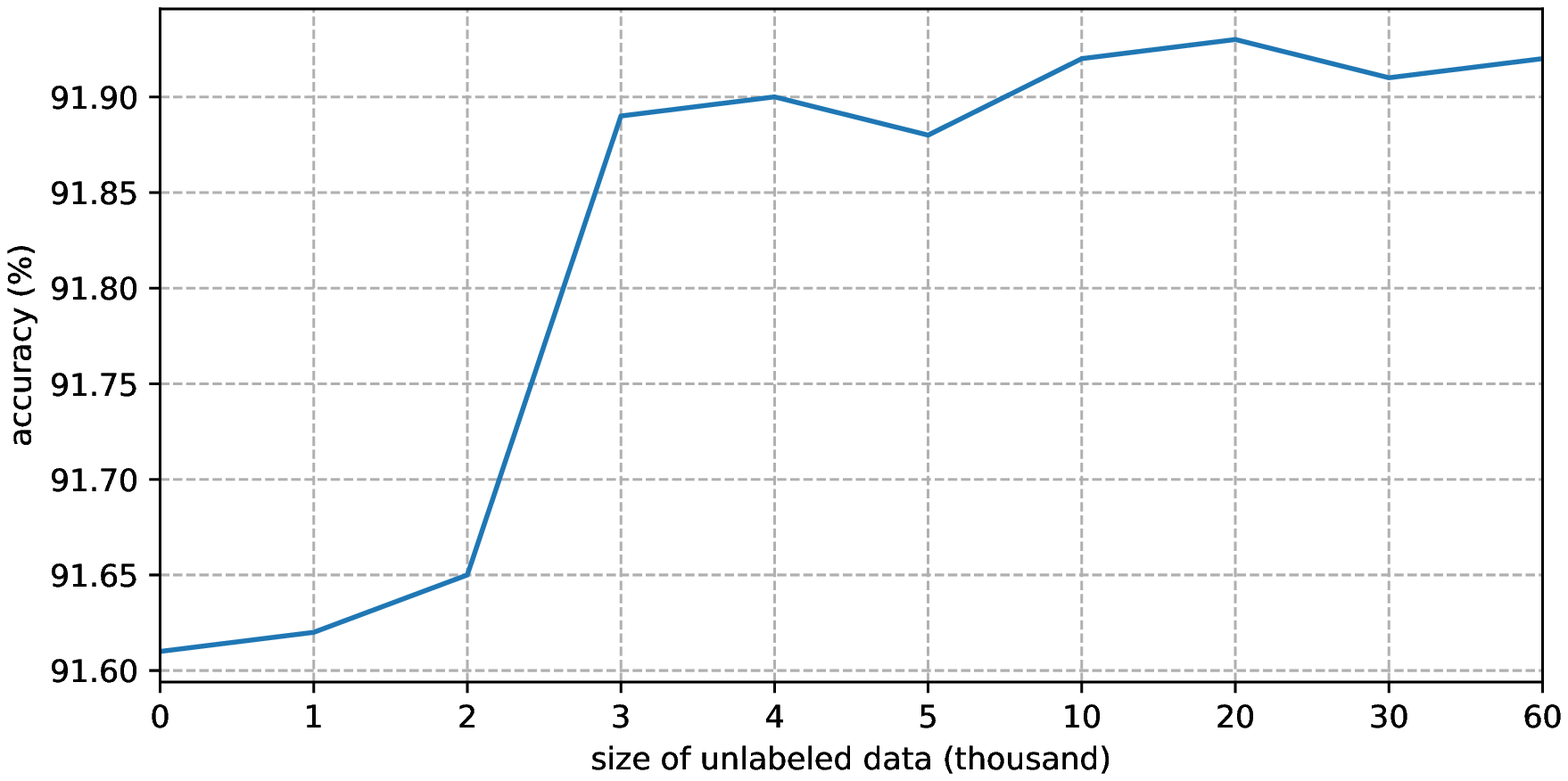}
\caption{Twitter dev accuracies (\%) when varying the amount of unlabeled data.}
 \label{fig:unlabeled}
\end{figure}
In order to examine the effect of unlabeled data, we report our Twitter dev accuracies when varying the unlabeled data size. We choose \vsmgghier as the model for this experiment since it benefits the most from unlabeled data. As Figure~\ref{fig:unlabeled} shows, gradually adding unlabeled data helps a little at the beginning. Further adding unlabeled data boosts the accuracy of the model.
The improvements that come from unlabeled data quickly plateau after the amount of unlabeled data goes beyond 10,000. This suggests that with little unlabeled data, the model is incapable of fully utilizing the information in the unlabeled data. However if the amount of unlabeled data is too large, the supervised training signal becomes too weak to extract something useful from the unlabeled data.

We also notice that when there is a large amount of unlabeled data, it is always better to pretrain the prior first using a small $\alpha$ (e.g., 0.1) and then use it as a warm start to train a new model using a larger $\alpha$ (e.g., 1.0).

Tuning the weight of the KL divergence could achieve a similar effect, but it may require tuning the weight for labeled data and unlabeled data separately. We prefer to pretrain the prior as it is simpler and involves less hyperparameter tuning.

\section{Conclusion}

We introduced variational sequential labelers for semi-supervised sequence labeling. They consist of  latent-variable generative models with flexible parametrizations for the variational posterior (using RNNs over the entire input sequence) and a classifier at each time step.  Our best models use multiple latent variables arranged in a hierarchical structure.

We demonstrate systematic improvements in NER and POS tagging accuracy across 8 datasets over a strong baseline.

We also find small, but consistent, improvements by using unlabeled data.

\section*{Acknowledgments}
We would like to thank NVIDIA for donating GPUs used in this research, the anonymous reviewers for their comments that improved this paper, and Google for a faculty research award to K.~Gimpel that partially supported this research.
This research was funded by NSF grant 1433485.

\bibliography{emnlp2018}
\bibliographystyle{acl_natbib_nourl}

\end{document}